\documentclass{article}

    \usepackage[preprint]{neurips_2024}

\usepackage[utf8]{inputenc} 
\usepackage[T1]{fontenc}    
\usepackage{hyperref}       
\usepackage{url}            
\usepackage{booktabs}       
\usepackage{amsfonts}       
\usepackage{nicefrac}       
\usepackage{microtype}      
\usepackage[table]{xcolor}         
\usepackage{natbib}
\setcitestyle{numbers,square}

\usepackage{amsmath}
\usepackage{amssymb}
\usepackage{mathtools}
\usepackage{amsthm}
\usepackage{algorithm}
\usepackage{algorithmic}
\usepackage{cleveref}
\usepackage{graphicx}
\usepackage{multirow}
\usepackage{subfigure}
\usepackage{threeparttable}
\usepackage{makecell}
\usepackage{color} 

\usepackage{wrapfig}

\title{Self-Distillation Learning Based on Temporal-Spatial Consistency for Spiking Neural Networks}

\author{
  Lin Zuo\thanks{Corresponding author.}, Yongqi Ding, Mengmeng Jing, Kunshan Yang, Yunqian Yu\\
  School of Information and Software Engineering\\
  University of Electronic Science and Technology of China\\
  \texttt{linzuo@uestc.edu.cn}\\ \texttt{\{yqding326,jingmeng1992,kinson199320,yuyunqianyyz\}@gmail.com}\\
}

\begin{document}

\maketitle

\begin{abstract}
  Spiking neural networks (SNNs) have attracted considerable attention for their event-driven, low-power characteristics and high biological interpretability. Inspired by knowledge distillation (KD), recent research has improved the performance of the SNN model with a pre-trained teacher model. However, additional teacher models require significant computational resources, and it is tedious to manually define the appropriate teacher network architecture. In this paper, we explore cost-effective self-distillation learning of SNNs to circumvent these concerns. Without an explicit defined teacher, the SNN generates pseudo-labels and learns consistency during training. On the one hand, we extend the timestep of the SNN during training to create an implicit temporal ``teacher" that guides the learning of the original ``student", i.e., the temporal self-distillation. On the other hand, we guide the output of the weak classifier at the intermediate stage by the final output of the SNN, i.e., the spatial self-distillation. Our temporal-spatial self-distillation (TSSD) learning method does not introduce any inference overhead and has excellent generalization ability. Extensive experiments on the static image datasets CIFAR10/100 and ImageNet as well as the neuromorphic datasets CIFAR10-DVS and DVS-Gesture validate the superior performance of the TSSD method. This paper presents a novel manner of fusing SNNs with KD, providing insights into high-performance SNN learning methods.
\end{abstract}

\section{Introduction}
\label{sec:intro}

Spiking neural networks (SNNs) model the information transmission mechanism of the biological neural system and transmit information through discrete spikes, yielding extremely low power consumption compared to artificial neural networks (ANNs)~\cite{ZUO20201}. In addition, the inherent temporal properties of spiking neurons enable SNNs with superior temporal feature extraction ability, making SNNs receive extensive attention from the research community~\cite{KDSNN, PALIF}.

Although SNNs show great promise, their training has always been plagued by the non-differentiability of spike activity. In order to obtain high-performance SNNs, some work pre-train an ANN and then convert it into an SNN with the same structure~\cite{9543525, offsetspike}. However, this practice corrupts the temporal feature extraction ability of SNNs, resulting in noticeable inference latencies~\cite{LTMD, 9423187}. Another feasible training method is the surrogate gradient learning. The gradient of non-differentiable spike activity is replaced with a smooth surrogate gradient during backpropagation, allowing SNNs to be trained using gradient descent~\cite{STBP,ponghiran_spiking_2022}. This practice achieves satisfactory performance even at low latencies, making it the most commonly used training method~\cite{Spikformer, ding2024shrinking, TKS}. Our work follows this line and further improves the performance of the SNN.

\begin{wrapfigure}{r}[0cm]{0pt}        
    \includegraphics[width=8cm]{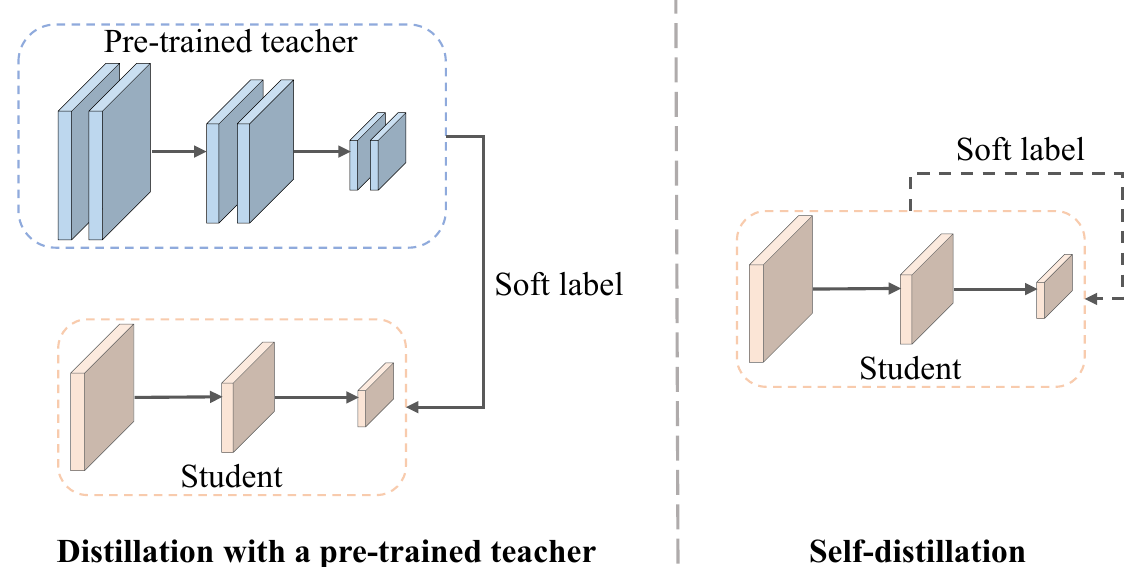}
    \caption{Comparison of distillation methods. Our self-distillation learning eliminates the need for the additional teacher model required for vanilla distillation, thus eliminating significant overhead.} 
    \label{fig:comdis}
\end{wrapfigure}

Recently, some work has introduced knowledge distillation (KD) to existing SNN training methods, typically using a large model with better performance to guide the learning of a lightweight model, to improve the performance of SNNs. As shown in Fig.~\ref{fig:comdis}, these distillation methods fall into two groups: (1) additional teacher models are required to guide the training of student SNN models~\cite{9412147,9410323,KDSNN,lv2023spikebert,bal2023spikingbert,hong2023lasnn,jointasnn}, and (2) without explicit teacher models, manually defined teacher or student SNN models generate guidance signals on their own~\cite{TKS,xu2023biologically}. For the first one, the teacher model incurs additional training time and memory overhead. Moreover, for satisfactory performance, the teacher model is typically an ANN~\cite{KDSNN,hong2023lasnn,jointasnn}, which does not improve the temporal feature extraction ability of the student SNN. Therefore, the architecture of the teacher model must be manually defined depending on the task and the student SNN. In contrast, for the second group,~\cite{xu2023biologically} uses a predefined fixed teacher signal, and~\cite{TKS} uses the output of the correct timesteps to guide the remaining timesteps, significantly reducing resource consumption. However, the fixed guidance signal in~\cite{xu2023biologically} lacks flexibility, and~\cite{TKS} is not available at very low timesteps (e.g., 1). More efficient and effective distillation learning strategies for SNNs still need to be further explored.

In this paper, starting from the inherent temporal and spatial properties of SNNs, we propose the temporal-spatial self-distillation (TSSD) learning method for SNNs. By extending the training timestep, TSSD considers the SNN with the extended timestep as the ``teacher" that guides the learning of the original small timestep ``student". The ``teacher" shares the same architecture and parameters as the ``student", without additional memory or computational overhead, and is continuously optimized during training to provide dynamic guidance to the ``student". In addition, this temporal-distillation decouples the timesteps for training and inference, allowing satisfactory inference performance at very low timesteps. On the other hand, during training, TSSD performs spatial self-distillation by adding a weak classifier in the intermediate stage of the SNN. The weak classifier makes predictions based on the features extracted in the intermediate stage and is guided by the final output of the SNN. This pushes the earlier stage of the SNN to extract features that are consistent with the whole network, thereby enhancing the feature extraction ability of the SNN. The weak classifier is discarded after training, therefore inference efficiency is not affected. In addition, our TSSD method is orthogonal to existing other methods such as various surrogate gradients, SNN architectures, and spiking neuron models, with superior generalizability. To evaluate the performance of the TSSD method, we perform extensive experiments on both static and neuromorphic datasets. The main contributions can be summarized as follows:
\begin{itemize}
\item We propose the TSSD learning method, which explores efficient self-distillation from the inherent temporal and spatial perspectives of SNNs, boosting performance without increasing inference overhead.
\item The TSSD method is orthogonal to other existing methods such as surrogate gradients, network architecture, and spiking neurons, and and can be seamlessly integrated, providing superior generalizability.
\item Extensive experiments on the both static and neuromorphic datasets validate the performance and generalizability of the TSSD method.
\end{itemize}

\section{Related Work}

\textbf{Surrogate Gradient Learning in SNNs.} Surrogate gradient-based learning method utilizes predefined smooth surrogate gradients during backpropagation, thus avoiding the problem of non-differentiable spike activity~\cite{STBP,8891809}. A large amount of work training deep SNNs based on surrogate gradients, such as efficient training methods~\cite{TET,IMLoss,RecDis} and normalization methods~\cite{tdBN,TEBN,MPBN}. Some works have designed novel SNN structures and trained them based on surrogate gradients~\cite{Spikformer,li2023efficient,autosnn}. Based on surrogate gradients, more efficient and biologically consistent spiking neuron models are also being explored~\cite{PLIF,MLF,BackEISNN,GLIF}. Our proposed TSSD method is based on the surrogate gradient method and is decoupled from specific network structures, spiking neuron models and surrogate gradient functions with superior generalizability.

\textbf{Knowledge Distillation.} Knowledge distillation (KD) defines a cumbersome teacher model and uses it to guide the training of a lightweight student model~\cite{hinton2015distilling}. According to the guiding information, KD can be categorized into logit distillation~\cite{MLD,li2022curriculum,Yuan_2020_CVPR,yuan2023studentfriendly,yang2023knowledge} and feature distillation~\cite{zong2023better,liu2023norm,romero2015fitnets,Heo_2019_ICCV}. The logit distillation guides the student to generate similar output logits with the teacher, while the feature distillation encourages the student to extract similar intermediate feature maps with the teacher. Both distillation methods require additional teacher models, and some self-distillation methods have achieved comparable results without explicit teachers~\cite{Yuan_2020_CVPR,Zipf,BYOT}. It is exactly the self-distillation learning in SNNs that we explored to improve performance while reducing distillation overhead.

\textbf{Distillation Learning in SNNs.} \cite{9412147} and~\cite{KDSNN} improved the performance of the student SNN by pre-training an ANN to guide the training of the SNN. \cite{9410323} and~\cite{10037455} first distill an ANN teacher and then convert the distilled ANN to SNN. \cite{jointasnn} and~\cite{hong2023lasnn} follow this path and optimize SNNs with a joint ANN. \cite{bal2023spikingbert} and~\cite{lv2023spikebert} combined this distillation mechanism to exploit SNNs for natural language processing, achieving excellent task performance. However, these methods requires pre-training of the teacher model, which imposes additional time and memory overhead. Another notable disadvantage is that ANN teachers limit the temporal feature extraction ability of SNN students and therefore can only work with static data~\cite{KDSNN,jointasnn}. \cite{xu2023biologically} avoids additional training overhead with manually defined teacher signals, but its suboptimal performance leaves room for improvement. \cite{TKS} divides the timestep into two parts based on the correctness of the generated prediction and guides the incorrect one with the correct output, achieving a remarkable performance. Nevertheless, \cite{TKS} does not work at very low timesteps, such as 1, when there is only a single prediction. In contrast to these methods, our method does not require pre-training of the teacher and only requires a slightly larger timestep and an additional weak classifier when training SNNs. Moreover, the trained SNN possesses superior spatio-temporal feature extraction ability and is capable of inference at ultra-low latency.

\section{Method}

\subsection{Spiking Neuron model}

SNNs are distinguished from ANNs by the transmission of information with 0-1 spikes and the temporal property stemming from spiking neurons. When a spiking neuron receives input current from presynaptic neurons, its membrane potential changes and generates spike to the next layer. In this paper, we use discrete leaky integrate-and-fire (LIF)~\cite{STBP} neurons, which mimic the properties of biological neurons with simplicity. Let the membrane potential of the $i$th neuron in layer $l$ be $H_{i}^{l}(t)$ and its response to the received current at timestep $t$ be:
\begin{equation}
H_{i}^{l}(t)=\left(1-\frac{1}{\tau}\right) H_{i}^{l}(t-1)+I_{i}^{l}(t),
\label{eq2}
\end{equation}
where $I_{i}^{l}(t)$ is the input current and $\tau$ is the leakage coefficient. After the membrane potential is updated, the spiking neuron calculates whether to generate a spike:
\begin{equation}
S_{i}^{l}(t) = \left\{
\begin{array}{cl}
1,\quad H_{i}^{l}(t) \ge \vartheta \\
0,\quad H_{i}^{l}(t) < \vartheta \\
\end{array},
\right.
\label{eq3}
\end{equation}
where $\vartheta$ is the firing threshold. The membrane potential is reset after the spike is generated. In this paper, we use the soft reset to reduce the membrane potential by the magnitude of the threshold:
\begin{equation}
H_{i}^{l}(t) = H_{i}^{l}(t)-S_{i}^{l}(t)\vartheta.
\label{eq4}
\end{equation}

The 0-1 spike of Eq.~\ref{eq3} is not differentiable, and therefore SNNs cannot be optimized directly by gradient descent. To train the SNN, we use the rectangular surrogate gradient~\cite{STBP} method to calculate the gradient of the spike $S_{i}^{l}(t)$ w.r.t. the membrane potential $H_{i}^{l}(t)$:
\begin{equation}
\frac{\partial S_{i}^{l}(t)}{\partial H_{i}^{l}(t)} \approx \frac{\partial h(H_{i}^{l}(t), \vartheta)}{\partial H_{i}^{l}(t)} = \frac{1}{a} \text{sign} (|H_{i}^{l}(t)-\vartheta|<\frac{a}{2}),
\label{eq5}
\end{equation}
where $a$ is the hyperparameter controlling the shape of the surrogate gradient function, which we set to 1.0.

\begin{figure*}[t]
  \centering
   \includegraphics[width=0.9\linewidth]{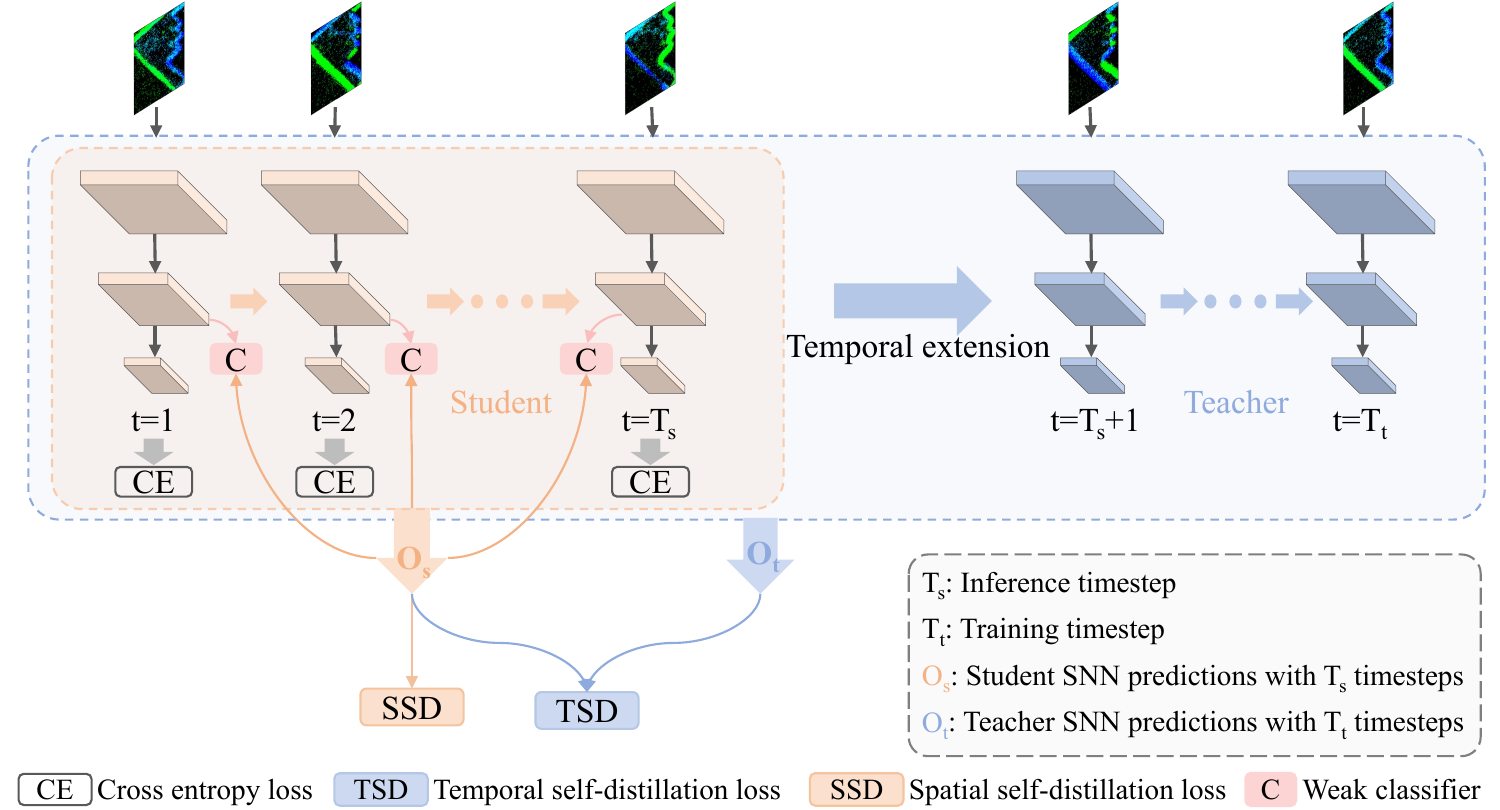}
   \caption{Illustration of the TSSD method. $T_s$ is the inference timestep, the training timestep is extended to $T_t>T_s$. The SNN with $T_t$ timesteps as the ``teacher" guides the training of the ``student" SNN with the top $T_s$ timesteps. Inference takes only $T_s$ timesteps with no additional overhead. From the spatial perspective, the final output of the SNN serves as the teacher to guide the weak output produced by the weak classifier. The weak classifier is discarded during inference.}
   \label{TSSD}
\end{figure*}

\subsection{Temporal Self-Distillation}

Spiking neurons iteratively accumulate membrane potential, fire spikes, and reset membrane potential over multiple timesteps $T$. Within a certain range, the larger the timestep $T$, the better the performance of the SNN~\cite{STBP,TET}. Inspired by this, we extend the temporal dimension of the SNN during training, with larger timesteps leading to better performance, thus yielding a ``teacher". The SNN before temporal extension serves as the ``student" and is guided by the ``teacher" during training for temporal self-distillation (TSD) learning, as shown in Fig.~\ref{TSSD}. In the following, we describe in detail this TSD learning method.

Let $f(\theta,T)$ denote an SNN, where $\theta$ is the network parameter and $T$ is the simulation timestep. Train the SNN to make $f(x; \theta,T) = y$, where $x$ is the input data and $y$ is the objective of the task (for the classification task, $y$ is the class label). For input $x$, $f(\theta,T)$ generates outputs at all $T$ timesteps, and the final decision $f(x; \theta,T)=\frac{\sum^T_{t=1}{f(x;\theta)}}{T}$ is considered as the average output over $T$ timesteps. The training process can be formulated as follows:
\begin{equation}
\min _{\theta} \sum_{x \in D}\mathcal{L}_{\text {task }}\left(f\left(x ; \theta,T\right), y\right),
\label{eq7}
\end{equation}
where $D$ is the training dataset and $L_{\text {task}}$ is the loss function, which is set to cross-entropy (CE) for the classification task.

For generalization, we denote two different timesteps by $T_s<T_t$. As a result, for the same set of parameters $\theta$, we obtain two logically different SNNs (with different timesteps): $f(\theta,T_s)$ and $f(\theta,T_t)$. Compared to $f(\theta,T_s)$, $f(\theta,T_t)$ works with larger timesteps, and ought to yield better performance. Therefore, we assign $f(\theta,T_t)$ as the ``teacher" to guide $f(\theta,T_s)$, the ``student" model, for TSD learning. This impels the ``student" to imitate the predictions generated by the ``teacher" for the same input, thus making more consistent decisions and improving the overall performance of the ``student" model. This process of TSD-guided learning can be formulated as:
\begin{equation}
\min _{\theta} \sum_{x \in D}\mathcal{L}_{\text{tsd}}\left(f\left(x ; \theta,T_s\right), f\left(x ; \theta,T_t\right)\right),
\label{eq8}
\end{equation}
where $L_{\text{tsd}}$ is the loss function for TSD learning, and in this paper we use the L2 distance for the distillation loss:
\begin{equation}
\left\|f\left(x ; \theta,T_s\right)-f\left(x ; \theta,T_t\right)\right\|_{2}^{2}.
\label{eq9}
\end{equation}

It is worth noting that the ``teacher" model shares the same parameter $\theta$ with the ``student" model. Thus there is no need to train additional teacher model as in~\cite{9412147,KDSNN}, only the timestep needs to be extended from $T_s$ to $T_t$ during training, while the $T_s$ timestep is still used for inference. This decouples the timesteps for training and inference, making inference feasible at ultra-low latencies (e.g., a timestep of 1). The settings for $T_t$ and $T_s$ should take into account the latency-performance tradeoffs during training and inference, which we will explore in~\cref{Experiments}. Another significant advantage is that both the ``teacher" and the ``student" are SNNs, thus the temporal feature extraction ability of the ``student" model is enhanced, as opposed to~\cite{KDSNN,jointasnn} where the ANN teacher causes the loss of the temporal feature extraction ability of the SNN student.

Combining Eq.~\ref{eq7} and Eq.~\ref{eq8}, the total loss for training SNNs based on the TSD can be formulated as:
\begin{equation}
\mathcal{L}_{\text{total}} = \mathcal{L}_{\text{task}} + \alpha \cdot \mathcal{L}_{\text{tsd}},
\label{eq10}
\end{equation}
where $\alpha$ is the coefficient controlling the weight of TSD.

\subsection{Spatial Self-Distillation}
In addition to the temporal property unique to SNNs, the spatial property as a common feature of deep neural networks has been used to explore distillation learning in ANNs~\cite{BYOT}. Here, we explore the spatial self-distillation (SSD) learning in SNNs for superior distillation performance. 

Typically, a neural network consists of many layers or blocks stacked on top of each other, such as the VGG~\cite{VGG} and the ResNet~\cite{ResNet} architecture. The earlier blocks extract the primary features and the later blocks extract the high-level abstract features for decision making. Taking this into account, we insert a weak classifier into the intermediate stage of the SNN during training to make predictions based on the extracted primary features. Compared to the complete SNN, the weak classifier is considered a smaller subnetwork, so the weak classifier and the complete SNN can be considered as a weak ``student" and a strong ``teacher". Our SSD method encourages the weak classifier to learn as much as possible the consistent output as the complete SNN, thereby contributing to the feature learning ability of the previous stages. The SSD learning is shown in Fig.~\ref{TSSD} and elaborated as follows.

Without loss of generality, assume that the weak classifier $C$ located in the middle of the SNN separates the SNN into two parts, $f(\theta)=f(\theta_1 \circ \theta_2)$. Within $T_s$ timesteps, the weak classifier generates an output $f(\theta_1 \circ C)$ based on $f(\theta_1)$ at each timestep, and these outputs are guided to align with the complete SNN average output:
\begin{equation}
\min _{\theta_1} \sum_{x \in D} \sum_{t \in T_s}\mathcal{L}_{\text{ssd}}\left(f_t\left(x ; \theta_1\circ C\right), f\left(x ; \theta,T_s\right)\right),
\label{eq11}
\end{equation}
where $f_t\left(x ; \theta_1\circ C\right)$ denotes the output of the weak classifier at timestep $t$, while $f\left(x ; \theta,T_s\right)$ represents the average output of the complete SNN over $T_s$ timesteps. This is similar to~\cite{BYOT}, where several additional bottlenecks are inserted into the ANN for distillation, but we further consider the multiple timestep output characteristics of the SNN. This allows the stable average output of the complete SNN to guide the unstable timestep-wise output of the weak classifier, further facilitating the learning of the previous stages.

As in Eq.~\ref{eq9}, we keep the L2 distance as the loss for SSD. This encourages logit matching of "student" and "teacher" , and also eliminates the need for tedious distillation temperature adjustment in the vanilla KL divergence loss~\cite{ijcai2021p362}. The total loss of training an SNN with SSD can be formulated as:
\begin{equation}
\mathcal{L}_{\text{total}} = \mathcal{L}_{\text{task}} + \beta \cdot \mathcal{L}_{\text{ssd}},
\label{eq12}
\end{equation}
where $\beta$ is the coefficient controlling the weight of SSD.

In this paper, the weak classifier consists of convolution, Batch Normalization (BN)~\cite{BN}, LIF neurons, and a fully connected layer, which requires only negligible overhead compared to the complete SNN. Note that the weak classifier is not used in the inference phase and therefore does not affect inference efficiency. Alternatively, if using the weak classifier to recognize simple samples during inference, it can achieve spatial early exit~\cite{BranchyNet} and shorten the forward propagation path to further improve the inference efficiency, which we explore in~\ref{earlyexit}.

\subsection{Temporal-Spatial Self-Distillation Learning}
With the proposed TSD and SSD methods, we can derive a joint TSSD method for training high-performance SNNs. The training framework is shown in Algo.~\ref{alg1}.

During training, the timestep was extended from $T_s$ to $T_t$ to generate a temporal ``teacher". For the input $x$, the previous stage of the SNN first generates the primary feature $f(x ; \theta_1,T_t)$, and then the later stage makes further predictions based on the primary feature. For the final prediction, ``teacher" $f(x ; \theta,T_t)$ guides ``student" $f(x ; \theta,T_s)$ in the temporal dimension to make a similar and stable prediction. On the other hand, the weak classifier generates a weak prediction $f(x ; \theta_1 \circ C,T_s)$ based on the primary feature $f(x ; \theta_1,T_s)$ and is guided by the stronger final prediction $f(x ; \theta,T_s)$. Note that the SSD works with timestep $T_s$, hence the weak classifier generates $T_s$ different outputs, which are individually guided by the final prediction.

The total loss of the TSSD method consists of task loss, TSD loss, and SSD loss:
\begin{equation}
\mathcal{L}_{\text{total}} = \mathcal{L}_{\text{task}} + \alpha \cdot \mathcal{L}_{\text{tsd}} + \beta \cdot \mathcal{L}_{\text{ssd}},
\label{eq13}
\end{equation}
\begin{equation}
\mathcal{L}_{\text{task}} = \sum_{x \in D} \sum_{t \in T_s} \text{Cross-Entropy} (f_t(x ; \theta), y).
\label{eq14}
\end{equation}

\begin{algorithm}[tb]
\caption{TSSD learning method for training SNNs} 
\label{alg1}
\begin{algorithmic}[1]
\REQUIRE Initialized SNN $f(\theta)$ and weak classifier $C$, training dataset $D$ with sample $x$ and label $y$, inference timestep $T_s$, training timestep $T_t$.
\ENSURE SNN trained by TSSD learning method.
\STATE \textit{\# Forward propagation.}

\STATE Generate the output $f(x ; \theta_1,T_s)$ of the intermediate stage over $T_s$ timesteps (\textbf{primary feature});
\STATE Generate the output $f(x; \theta_1 \circ C,T_s)$ of the weak classifier over $T_s$ timesteps (\textbf{spatial student});
\STATE Generate the final output $f(x;\theta,T_s)$ (\textbf{spatial teacher}) and the output $f(x;\theta,T_t)$ with extended timestep (\textbf{temporal teacher});
\STATE $\mathcal{L}_{\text{tsd}} \leftarrow$Eq.~\ref{eq8};\textit{\# Temporal self-distillation loss.}
\STATE $\mathcal{L}_{\text{ssd}} \leftarrow$Eq.~\ref{eq11};\textit{\# Spatial self-distillation loss.}
\STATE $\mathcal{L}_{\text{task}} \leftarrow$Eq.~\ref{eq14}; \textit{\# Task loss.}
\STATE $\mathcal{L}_{\text{total}} \leftarrow$Eq.~\ref{eq13}; \textit{\# Total loss.}
\STATE \textit{\# Back propagation.}
\STATE Calculate the gradient $\frac{\partial \mathcal{L}_{\text{total}}}{\partial \theta}$ and $\frac{\partial \mathcal{L}_{\text{total}}}{\partial C}$ ;
\STATE Update $\theta$ and $C$.
\end{algorithmic}
\end{algorithm}

\vspace{-0.4cm}
\subsection{Theoretical Analysis}
We theoretically analyze the TSSD method from the perspective of empirical risk minimization. Given a training sample $S={\{(x_n,y_n)\}_{n=1}^N} \sim \mathbb{P}^N$, we train the SNN $f(\theta)$ with minimal risk $R(f)= \underset{(x,y)\sim \mathbb{P}}{\mathbb{E}}[\mathcal{L}(f(x),y)]$. For one-hot encoded label $e_y \in (0,1)^{L}$, the risk $R(f)$ is approximated by \textit{empirical risk}:
\begin{equation}
\hat{R}(f;S) = \frac{1}{N}\sum_{n\in[N]}e_{yn}^{\top}\mathcal{L}(f(x_n)).
\label{eq15}
\end{equation}

In \cite{menon21a}, an elementary observation for the population risk $R(f)$ is:
\begin{equation}
R(f) = \underset{x}{\mathbb{E}}[\underset{y|x}{\mathbb{E}}[\mathcal{L}(f(x),y)]]=\underset{x}{\mathbb{E}}[p^{*}(x)^{\top}\mathcal{L}(f(x))],
\label{eq16}
\end{equation}
where $p^{*}(x)=[ \mathbb{P} (y|x)]_{y \in [L]}$ is the \textit{Bayes class probability distribution} with intrinsic confusion across multiple labels without concentrating on a specific label, similar to our non-deterministic teacher output. Taking it further, the \textit{Bayes-distilled risk} of the sample $S\sim \mathbb{P}^N$ can be defined as:
\begin{equation}
\hat{R}_{*}(f;S) = \frac{1}{N}\sum_{n\in[N]}p^{*}(x_n)^{\top}\mathcal{L}(f(x_n)).
\label{eq17}
\end{equation}
According to \cite{menon21a}, for discriminative predictors and non-deterministic labels, the Bayes-distilled risk $\hat{R}_{*}(f;S)$ has a much lower variance than the normal risk $\hat{R}(f;S)$. For our TSSD method, the teacher outputs in both temporal and spatial dimensions can be considered as non-deterministic soft labels. Therefore, our student SNN model learns this non-deterministic labels and is able to have lower Bayes-distilled risk, i.e., lower variance and higher performance.

From another perspective, the teacher has a relatively more stable output and lower variance than the student, which benefits the student in learning to represent information, consistent with~\cite{Wang_2022_CVPR}.

\section{Experiments}
\label{Experiments}

\subsection{Implementation Details}

We evaluate our method on the static CIFAR10/100~\cite{CIFAR100}, ImageNet~\cite{ImageNet} and the neuromorphic datasets CIFAR10-DVS~\cite{CIFAR10-DVS} and DVS-Gesture~\cite{DVS-Gesture}. Two different types of network architectures, VGG and ResNet, are used in the experiments. We report the average accuracy and standard deviation of three experiments. The detailed experimental setup can be found in the Appendix \ref{ExperimentsDetail}.

\vspace{-0.2cm}
\subsection{Ablation Study}
\label{Ablation}

\begin{wraptable}{r}{7.3cm}
\tabcolsep=0.008\columnwidth
  \caption{Ablation studies for the TSSD method.}
   \scalebox{0.84}{
  \begin{tabular}{ccccc}
    \toprule
    \multirow{2}{*}{Dataset} & \multirow{2}{*}{Method} & \multirow{2}{*}{$T_s$} & \multicolumn{2}{c}{Acc$\pm$std (\%)} \\
    & & & VGG & ResNet\\ 
    \midrule
    \multirow{4}{*}{CIFAR10}
    & Baseline & 2 & 93.35$\pm$0.12 & 92.85$\pm$0.35\\
    & +TSD & 2 & 93.99$\pm$0.05 & 93.42$\pm$0.13\\
    & +SSD & 2 & 93.90$\pm$0.08 & 93.14$\pm$0.14\\
    & +TSSD & 2 & \textbf{94.41}$\pm$0.05 & \textbf{93.48}$\pm$0.02\\
    \hline
    \multirow{4}{*}{CIFAR10-DVS}
    & Baseline & 5 & 66.87$\pm$0.05 & 58.60$\pm$0.54\\
    &+TSD & 5 & 70.93$\pm$0.87 & 59.13$\pm$0.33\\
    &+SSD & 5 & 70.63$\pm$0.21 & 60.70$\pm$0.57\\
    &+TSSD & 5 & \textbf{72.90}$\pm$0.37 & \textbf{63.57}$\pm$0.59\\
    \bottomrule
  \end{tabular}
  }
\label{tab:ablation}
\end{wraptable}

\vspace{-0.2cm}
\textbf{Comparison with the baseline SNN.} We explore the effectiveness of TSD and SSD learning by setting both the loss coefficients $\alpha$ and $\beta$ to 1.0. The experiments were performed on CIFAR10 and CIFAR10-DVS with $T_t=2T_s$. The experimental results are shown in Table~\ref{tab:ablation}. It can be seen that both TSD and SSD are able to improve the performance of the baseline, with TSSD maximizing the performance gains. Moreover, this consistent performance gain on both static and neuromorphic datasets suggests that the TSSD method enhances both the spatial and temporal feature extraction capabilities of SNNs, which is hard to achieve with additional ANN teachers. For further comparison, accuracy change curves and spike firing rate maps are provided in Appendix~\ref{acc} and Appendix~\ref{sfram}.

\begin{wrapfigure}{r}[0cm]{0pt}        
    \includegraphics[width=7.5cm]{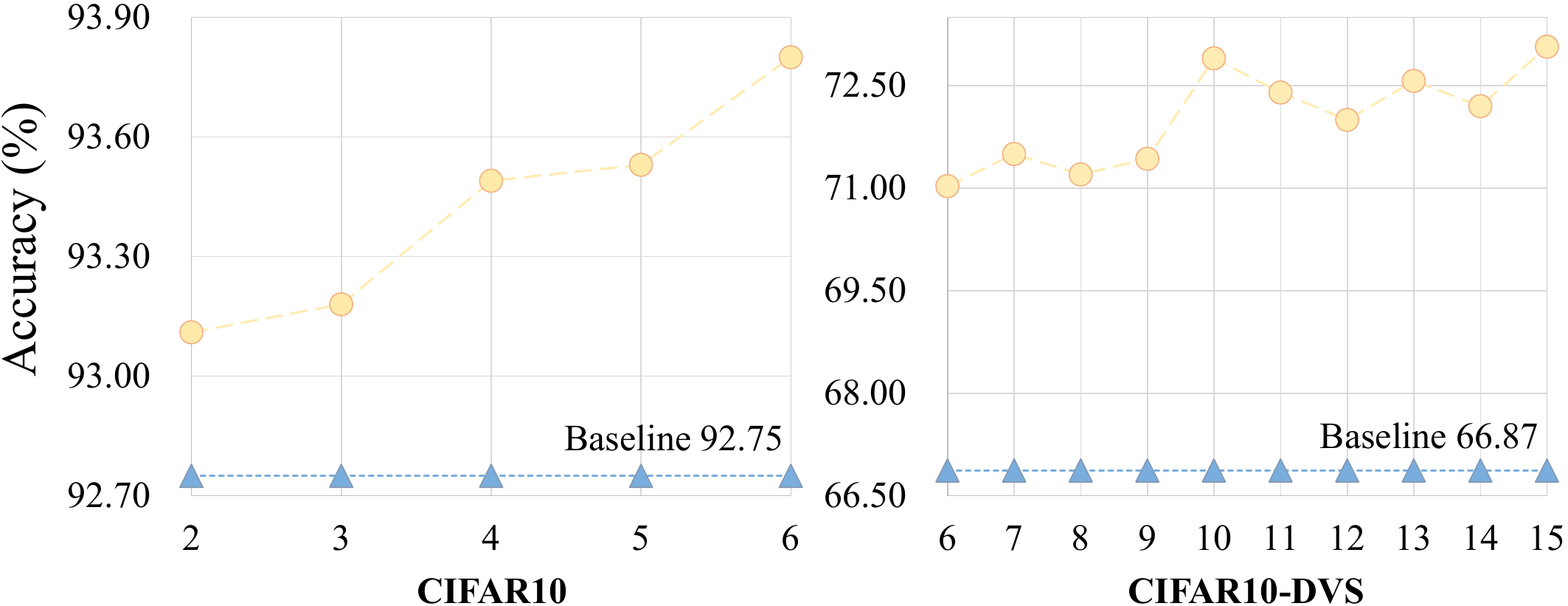}
    \caption{Influence of $T_t$. The overall performance of the TSSD method improves as $T_t$ increases.} 
    \label{fig:Tt}
\end{wrapfigure}
\textbf{Influence of $T_t$.} We investigate the influence of the training timestep on the performance, since $T_t$ has a direct impact on the temporal ``teacher'' model. $T_t$ ranges from [2,6], $T_s=1$ for CIFAR10, and ranges from [6,15], $T_s=5$ for CIFAR10-DVS. As shown in Fig.~\ref{fig:Tt}, for the fixed $T_s$, the larger $T_t$ gives better performance, although it fluctuates slightly on the CIFAR10-DVS. For any value of $T_t$, TSSD consistently outperforms the vanilla baseline. Even when $T_t$ was only 1 greater than $T_s$, the accuracy of TSSD was 0.36\% and 4.16\% greater than baseline, respectively. It is worth noting that an accuracy of 93.80\% can be achieved on CIFAR10 when $T_t=6$ and $T_s=1$, indicating that our method can achieve satisfactory performance with ultra-low latency.

\begin{table}[h]
\tabcolsep=0.011\columnwidth
  \centering
  \caption{Comparative results between TSSD and vanilla SNNs trained over $T_t$ timesteps for inference over $T_s$ timesteps.}
  \scalebox{0.84}{
  \begin{tabular}{c|cccc|cccc}
    \toprule
    \multirow{5}{*}{CIFAR10} & & $T_t$ & 6 & 4 &  & $T_t$ & 6 & 4 \\
    \cline{2-9}
    & \multirow{4}{*}{$T_s=1$} & Vanilla (VGG) & 35.67\% & 44.53\% & \multirow{4}{*}{$T_s=2$} & Vanilla (VGG) & 92.77\% & 93.20\%\\
    && \cellcolor{pink!20} TSSD (VGG) & \cellcolor{pink!20} \textbf{93.80\%} & \cellcolor{pink!20} \textbf{93.49\%} & & \cellcolor{pink!20}TSSD (VGG) & \cellcolor{pink!20}\textbf{94.30\%} & \cellcolor{pink!20}\textbf{94.41\%}\\
    \cline{3-5}\cline{7-9}
    && Vanilla (ResNet) & 39.25\% & 50.29\% & & Vanilla (ResNet) & 91.66\% & 92.65\%\\
    && \cellcolor{pink!20}TSSD (ResNet) & \cellcolor{pink!20}\textbf{92.27\%} & \cellcolor{pink!20}\textbf{92.25\%} & & \cellcolor{pink!20}TSSD (ResNet) & \cellcolor{pink!20}\textbf{93.66\%} & \cellcolor{pink!20}\textbf{93.48\%}\\
    \hline
     \multirow{5}{*}{CIFAR10-DVS} & & $T_t$ & 15 & 10 &  & $T_t$ & 15 & 10 \\
     \cline{2-9}
    & \multirow{4}{*}{$T_s=5$} & Vanilla (VGG) & 62.45\% & 66.07\% & \multirow{4}{*}{$T_s=7$} & Vanilla (VGG) & 69.67\% & 70.23\%\\
    && \cellcolor{pink!20}TSSD (VGG) & \cellcolor{pink!20}\textbf{73.20\%} & \cellcolor{pink!20}\textbf{72.90\%} & & \cellcolor{pink!20}TSSD (VGG) & \cellcolor{pink!20}\textbf{77.30\%} & \cellcolor{pink!20}\textbf{76.40\%}\\
    \cline{3-5}\cline{7-9}
    && Vanilla (ResNet) & 54.47\% & 57.27\% & & Vanilla (ResNet) & 60.80\% & 62.83\%\\
    && \cellcolor{pink!20}TSSD (ResNet) & \cellcolor{pink!20}\textbf{65.50\%} & \cellcolor{pink!20}\textbf{62.97\%} & & \cellcolor{pink!20}TSSD (ResNet) & \cellcolor{pink!20}\textbf{69.93\%} & \cellcolor{pink!20}\textbf{67.83\%}\\
    \bottomrule
  \end{tabular}
  }
  \label{tab:vanilla}
\end{table}

\begin{table*}[!tb]
 \centering
 \caption{Comparative results on static datasets (\%). * denotes self-implementation results.}
 \begin{threeparttable}
 \scalebox{0.84}{
 \begin{tabular}{cccccc}
  \toprule
 Method &Type & Architecture & $T$ & CIFAR10 & CIFAR100\\
  \midrule
   \multirow{2}{*}{RMP-Loss~\cite{RMPloss}} &\multirow{2}{*}{Surrogate gradient} &\multirow{2}{*}{VGG16} & 10 & 94.39 & 73.30 \\  &&& 4 & 93.33 & 72.55\\
   \cline{2-6}
   MLF \cite{MLF} & Surrogate gradient & DS-ResNet 20 & 4 & 94.25 & -\\
  Spikformer \cite{Spikformer} & Surrogate gradient & Spikformer-4-256 & 4 & 93.94 & -\\
  KDSNN~\cite{KDSNN} & Surrogate gradient+KD & ResNet-18 & 4 & 93.41 & -\\
  TET \cite{TET} & Surrogate gradient & ResNet-19 & 2 & 94.16 & 72.87\\
  Real Spike \cite{real_spike_2022} & Surrogate gradient & ResNet-19/VGG-16 & 2/5 & 94.01 & 70.62\\
  IM-Loss+ESG \cite{IMLoss} & Surrogate gradient & ResNet-19/VGG-16 & 2/5 & 93.85 & 70.18\\
  MPBN~\cite{MPBN} & Surrogate gradient & ResNet-20 & 2 & 93.54 & 70.79\\
  teacher default-KD~\cite{xu2023biologically} & Surrogate gradient+KD & VGG-9\tnote{*} & 2 & 93.49$\pm$0.03 & 74.31$\pm$0.09\\
  \cline{2-6}
  \multirow{2}{*}{SRP~\cite{SRP}} & \multirow{2}{*}{Conversion} & \multirow{2}{*}{VGG-16} & 2 & - & 74.31 \\ &&& 1 & - & 71.52\\
  \hline
  \multirow{3}{*}{\textbf{TSSD (Ours)}} &\multirow{3}{*}{Surrogate gradient+KD} & VGG-9 & 2 & \textbf{94.41}$\pm$0.05 & \textbf{74.69}$\pm$0.10\\&& VGG-9 & 1 & 93.49$\pm$0.17 & 73.33$\pm$0.12\\ && ResNet-18 & 2 & 93.37$\pm$0.09 & 73.40$\pm$0.17\\
  \bottomrule
 \end{tabular}
 }
 \end{threeparttable}
 \label{com_static}
\end{table*}

In addition, we compare vanilla SNNs trained over training timestep $T_t$ with our TSSD method for $T_s$ timesteps inference. As can be seen from Table~\ref{tab:vanilla}, when the inference timestep $T_s$ is smaller than the training timestep $T_t$, vanilla SNNs are far inferior to our TSSD method, especially on CIFAR10. The reason is that $T_t$ and $T_s$ influence the distribution of information, and this distribution difference affects the inference of vanilla SNNs, leading to poorer performance with larger timestep differences. Another reason for the performance degradation resonating with~\cite{ren2023spiking} is that the large timestep during training leads to a more severe problem of mismatch between the surrogate gradient and the true gradient. In contrast, our TSSD method simultaneously integrates the information of both $T_s$ and $T_t$ timesteps during training, and fully exploits the large training timestep while avoiding these degradation risks, thus improving the inference performance of SNNs.

\begin{wrapfigure}[13]{r}[0cm]{0pt}        
    \includegraphics[width=5.2cm]{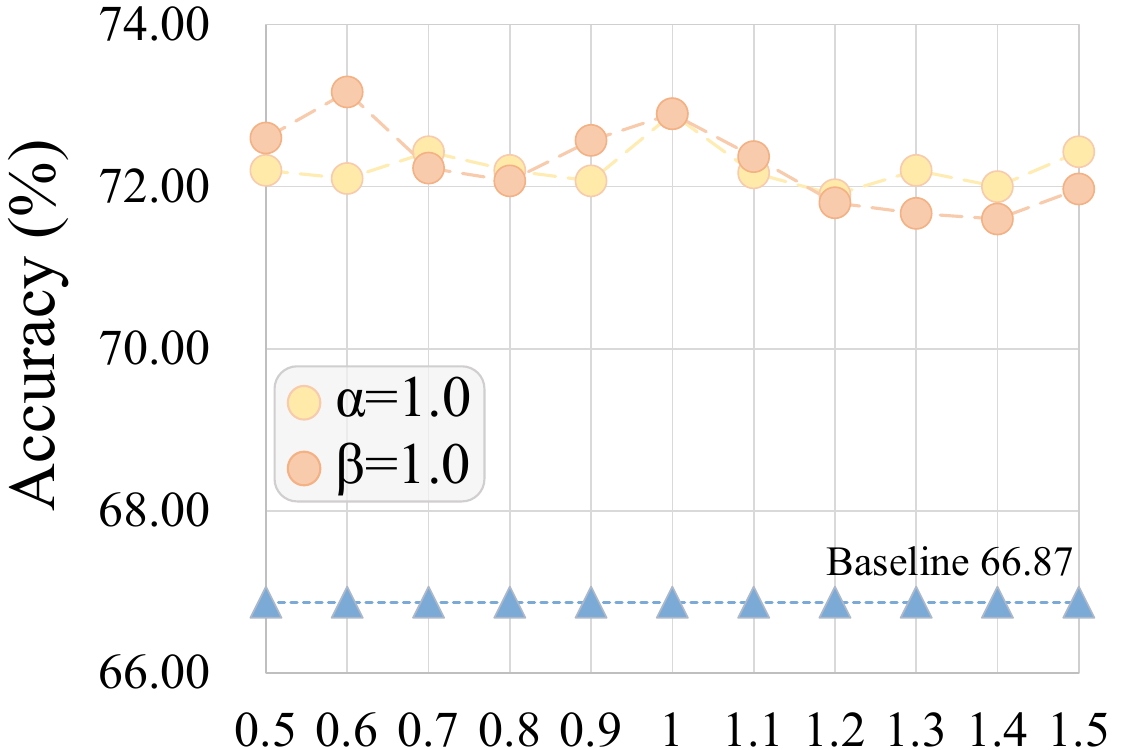}
    \caption{Influence of $\alpha$ and $\beta$. Our TSSD method is insensitive to $\alpha$ and $\beta$, and consistently yielding much better performance than the baseline.} 
    \label{fig:ab}
\end{wrapfigure}
\textbf{Influence of loss coefficients.} We explore the influence of $\alpha$ and $\beta$ on the TSSD method on CIFAR10-DVS. We fixed one of them to be 1.0 and the other to range from 0.5 to 1.5, and the results are shown in Fig.~\ref{fig:ab}. The experimental results show that our TSSD method is insensitive to $\alpha$ and $\beta$, and yields much higher accuracy than the baseline model for any value of $\alpha$ and $\beta$, although with slight fluctuations.

\subsection{Comparison with Other Methods}
\begin{table*}[!tb]
 \centering
 \tabcolsep=0.008\columnwidth
 \caption{Comparative results on neuromorphic datasets (\%), * denotes self-implementation results. $\dag$ indicates using data augmentation, and $T$ is the inference timestep.}
 \scalebox{0.84}{
 \begin{threeparttable}
 \begin{tabular}{cccccc}
  \toprule
  Method & Type & Architecture & $T$ & CIFAR10-DVS & DVS-Gesture\\
  \midrule
  DSR~\cite{DSR} & Differential spike & VGG-11\tnote{$\dag$} & 20 & 77.27 & -\\
  SDT~\cite{yao2024spike} & Surrogate gradient & Spike-driven Transformer & \cellcolor{blue!10}16 & \cellcolor{blue!10}80.00\tnote{$\dag$} & \cellcolor{blue!10}94.33$\pm$0.71\tnote{*} \\
  GLIF~\cite{GLIF} & Surrogate gradient & 7B-wideNet & \cellcolor{blue!10}16 & \cellcolor{blue!10}76.80 & -\\
  LSG~\cite{lian23learnable} & Learnable gradient & ResNet-19 & 10 & 77.90 & -\\
  TET~\cite{TET} & Surrogate gradient & VGGSNN & 10 & 77.33 & -\\
  MPBN~\cite{MPBN} & Surrogate gradient & ResNet-19 & 10 & 74.40 & - \\
  Real Spike~\cite{real_spike_2022} & Surrogate gradient & ResNet-19 & 10 & 72.85 & -\\
  \cline{2-6}
  \multirow{3}{*}{ASGL~\cite{ASGL}} & \multirow{3}{*}{Adaptive gradient} & \multirow{3}{*}{VGGSNN\tnote{*}} & \cellcolor{blue!10}16 & - & \cellcolor{blue!10}92.13$\pm$0.16 \\ &&& \cellcolor{green!8}8 & \cellcolor{green!8}76.33$\pm$0.12\tnote{$\dag$} & - \\ &&& \cellcolor{pink!20}5 & - &  \cellcolor{pink!20}79.28$\pm$0.16\\
  \cline{2-6}
 \multirow{2}{*}{MLF~\cite{MLF}} & \multirow{2}{*}{Surrogate gradient} & VGG-9\tnote{*} & \cellcolor{pink!20}5 & \cellcolor{yellow!10}67.07$\pm$0.56 & \cellcolor{pink!20}85.77$\pm$1.30 \\  && DS-ResNet20 & 40 & - & 97.29\\
 \cline{2-6}
 \multirow{2}{*}{PLIF~\cite{PLIF}} & \multirow{2}{*}{Surrogate gradient} & \multirow{2}{*}{VGG9\tnote{*}} & \cellcolor{pink!20}5 & \cellcolor{yellow!10}66.77$\pm$0.17 & \cellcolor{pink!20}79.98$\pm$0.82 \\ &&& \cellcolor{blue!10}16 & - & \cellcolor{blue!10}94.68$\pm$0.71\\
 \cline{2-6}
  \multirow{2}{*}{TEBN~\cite{TEBN}} & \multirow{2}{*}{Surrogate gradient} & \multirow{2}{*}{VGG9\tnote{*}} & \cellcolor{pink!20}5 & \cellcolor{yellow!10}67.70$\pm$3.55 & \cellcolor{pink!20}78.70$\pm$1.40 \\ &&& \cellcolor{blue!10}16 & - & \cellcolor{blue!10}94.56$\pm$0.16\\
   \cline{2-6}
   \multirow{2}{*}{Spikformer~\cite{Spikformer}} & \multirow{2}{*}{Surrogate gradient} & \multirow{2}{*}{Spikformer\tnote{*}} & \cellcolor{pink!20}5 & \cellcolor{yellow!10}68.55$\pm$1.39\tnote{$\dag$} & \cellcolor{pink!20}79.86$\pm$0.35\\ &&& \cellcolor{blue!10}16 & - & \cellcolor{blue!10}95.25$\pm$0.82\\
   \cline{2-6}
  \multirow{2}{*}{teacher default-KD~\cite{xu2023biologically}} & \multirow{2}{*}{Surrogate gradient+KD} & \multirow{2}{*}{VGG-9\tnote{*}} & \cellcolor{pink!20}5 & \cellcolor{yellow!10}66.37$\pm$0.41 & \cellcolor{pink!20}82.75$\pm$0.71\\  &&& \cellcolor{blue!10}16 & - & \cellcolor{blue!10}96.18$\pm$0.29\\
  \hline
  \multirow{5}{*}{\textbf{TSSD (Ours)}} & \multirow{5}{*}{Surrogate gradient+KD} & \multirow{3}{*}{VGG-9} & \cellcolor{pink!20}5 & \cellcolor{yellow!10}72.90$\pm$0.37 &  \cellcolor{pink!20}86.69$\pm$0.59 \\&&& \cellcolor{green!8}8 & \cellcolor{green!8}78.70$\pm$0.62 & \cellcolor{green!8}90.85$\pm$0.99\\ &&& \cellcolor{blue!10}16 & \cellcolor{blue!10}\textbf{84.37}$\pm$0.52 & \cellcolor{blue!10}\textbf{97.45}$\pm$0.16\\\cline{3-6} && \multirow{2}{*}{ResNet-18} & \cellcolor{green!8}8 & \cellcolor{green!8}72.90$\pm$0.33 & \cellcolor{green!8}87.96$\pm$0.66\\ &&& \cellcolor{blue!10}16 & \cellcolor{blue!10}81.60$\pm$0.67 & \cellcolor{blue!10}95.60$\pm$0.16\\
  \bottomrule
 \end{tabular}
 \end{threeparttable}
 }
 \label{com_dvs}
\end{table*}

We compare TSSD with other existing methods on static and neuromorphic datasets. For self-implemented methods, implementation details are provided in Appendix \ref{self-implemented}.

\textbf{Static datasets.} For static datasets, we set $T_t=4$ and the comparative results on CIFAR10/100 are shown in Table~\ref{com_static}. At the same or lower timestep, our method achieves better performance than the comparative methods. In particular, even with only 1 timestep, our VGG-9 achieves an average test accuracy of 73.33\% on CIFAR100, which even surpasses the performance of RMP-Loss~\cite{RMPloss} at 10 timesteps. The comparative results on ImageNet are shown in Table~\ref{com_imagenet}. Our ResNet-34 achieves an accuracy of 66.13\% with two timesteps, outperforming other comparative methods at the same latency. Even when compared with TET~\cite{TET}, MPBN~\cite{MPBN}, and RMP-Loss~\cite{RMPloss}, our model achieves higher accuracy with lower latency. This confirms the effectiveness of our method on large-scale challenging datasets.

\textbf{Neuromorphic datasets.} For neuromorphic datasets, we set $T_t=2T_s$ and the experimental results are shown in Table~\ref{com_dvs}. For CIFAR10-DVS, our VGG-9 achieves an average test accuracy of 78.70\% at 8 timesteps, which even exceeds the accuracy of DSR~\cite{DSR} and TET~\cite{TET} at 20 and 10 timesteps, respectively. Even with only 5 timesteps, our method achieves an average test accuracy of 72.90\%, surpassing MLF~\cite{MLF}, TEBN~\cite{TEBN}, teacher default-KD~\cite{xu2023biologically}, and Spikformer~\cite{Spikformer} under the same conditions, and even Real Spike~\cite{real_spike_2022} with 10 timesteps. For DVS-Gesture, TSSD outperforms the comparative models both at 16 timesteps and at low latencies of only 5 timesteps. In particular, our VGG-9 at 16 timesteps is 0.16\% more accurate than MLF~\cite{MLF} at 40 timesteps. Experiments on neuromorphic datasets strongly confirm the effectiveness of our method to improve the temporal feature extraction capability of SNNs.

\vspace{-0.1cm}
\subsection{Further Extension and Generalization}
Here we explore further extensions to TSSD and compatibility with existing methods. Guiding the weak classifier of the intermediate stage with the final output over $T_t$ timesteps enables a natural extension of the TSSD method. We evaluate the performance gains of this extension in Table~\ref{TSSDextension}. The experimental results show that the average test accuracy on CIFAR10-DVS and DVS-Gesture is increased by 0.43\% and 1.39\%, respectively. In addition, our TSSD method can be integrated with various surrogate gradient functions~\cite{lian23learnable,9746774}, finer-grained spiking neurons~\cite{PLIF,GLIF,MLF}, and BN dedicated to SNNs~\cite{tdBN,TEBN,MPBN}, which can dramatically improve the performance of SNNs. For instance, TSSD extension experiments for ASGL~\cite{ASGL}, MLF~\cite{MLF}, and TEBN~\cite{TEBN} are shown in Table~\ref{TSSDextension}. The accuracy of these methods for neuromorphic object recognition is greatly improved, demonstrating the generalizability of TSSD. We consider more extensions as future work.

\begin{table}[h]
\tabcolsep=0.005\columnwidth
  \centering
  \caption{Extension of the TSSD method to existing methods and further extension (\%).}
  \scalebox{0.84}{
  \begin{tabular}{c|ccc|ccc|ccc|ccc}
    \toprule
    & ASGL & & +TSSD & MLF & & +TSSD & TEBN & & +TSSD & TSSD & & Further extension\\
    \hline
    CIFAR10-DVS & 76.33 & $\rightarrow$ & $79.43_{\textcolor{red!15!green}{+3.10}}$ & 67.07& $\rightarrow$ & $73.03_{\textcolor{red!15!green}{+5.96}}$ & 67.70& $\rightarrow$ & $70.27_{\textcolor{red!15!green}{+2.57}}$ & 72.90 & $\rightarrow$ & $73.33_{\textcolor{red!15!green}{+0.43}}$\\
    DVS-Gesture & 92.13 & $\rightarrow$ & $94.67_{\textcolor{red!15!green}{+2.54}}$ & 85.77 & $\rightarrow$ & $89.59_{\textcolor{red!15!green}{+3.82}}$ & 78.70 & $\rightarrow$ & $81.25_{\textcolor{red!15!green}{+2.55}}$ & 86.69 & $\rightarrow$ & $88.08_{\textcolor{red!15!green}{+1.39}}$\\
    \bottomrule
  \end{tabular}
  }
  \label{TSSDextension}
\end{table}

\section{Conclusion}
\label{sec:conclusion}

This paper explores self-distillation learning in SNNs to alleviate the heavy overhead and inefficiency challenges of traditional KD, while enhancing the performance of SNNs. To this end, we propose the TSSD learning method, which self-distills the SNN in both temporal and spatial perspectives and guides the SNN to learn consistency. Extensive experiments on both static and neuromorphic datasets showed consistent performance gains, demonstrating the effectiveness of our method. 

A limitation of the TSSD method is that it introduces additional training overhead, but this is modest compared to other distillation methods that require heavy teacher models. In addition, TSSD is compatible with other SNN architectures, spiking neurons, and surrogate gradient methods, leaving a wide scope for extension. We expect this work will contribute to high-performance and efficient SNNs.

{
\small
\bibliographystyle{unsrt}
\bibliography{main}

\begin{thebibliography}{10}

\bibitem{ZUO20201}
Lin Zuo, Yi~Chen, Lei Zhang, and Changle Chen.
\newblock A spiking neural network with probability information transmission.
\newblock {\em Neurocomputing}, 408:1--12, 2020.

\bibitem{KDSNN}
Qi~Xu, Yaxin Li, Jiangrong Shen, Jian~K. Liu, Huajin Tang, and Gang Pan.
\newblock Constructing deep spiking neural networks from artificial neural networks with knowledge distillation.
\newblock In {\em Proceedings of the IEEE/CVF Conference on Computer Vision and Pattern Recognition (CVPR)}, pages 7886--7895, June 2023.

\bibitem{PALIF}
Yongqi Ding, Lin Zuo, Kunshan Yang, Zhongshu Chen, Jian Hu, and Tangfan Xiahou.
\newblock An improved probabilistic spiking neural network with enhanced discriminative ability.
\newblock {\em Knowledge-Based Systems}, 280:111024, 2023.

\bibitem{9543525}
Jibin Wu, Chenglin Xu, Xiao Han, Daquan Zhou, Malu Zhang, Haizhou Li, and Kay~Chen Tan.
\newblock Progressive tandem learning for pattern recognition with deep spiking neural networks.
\newblock {\em IEEE Transactions on Pattern Analysis and Machine Intelligence}, 44(11):7824--7840, 2022.

\bibitem{offsetspike}
Zecheng Hao, Jianhao Ding, Tong Bu, Tiejun Huang, and Zhaofei Yu.
\newblock Bridging the gap between {ANN}s and {SNN}s by calibrating offset spikes.
\newblock In {\em The Eleventh International Conference on Learning Representations}, 2023.

\bibitem{LTMD}
Siqi Wang, Tee~Hiang Cheng, and Meng-Hiot Lim.
\newblock Ltmd: Learning improvement of spiking neural networks with learnable thresholding neurons and moderate dropout.
\newblock In S.~Koyejo, S.~Mohamed, A.~Agarwal, D.~Belgrave, K.~Cho, and A.~Oh, editors, {\em Advances in Neural Information Processing Systems}, volume~35, pages 28350--28362. Curran Associates, Inc., 2022.

\bibitem{9423187}
Souvik Kundu, Gourav Datta, Massoud Pedram, and Peter~A. Beerel.
\newblock Spike-thrift: Towards energy-efficient deep spiking neural networks by limiting spiking activity via attention-guided compression.
\newblock In {\em 2021 IEEE Winter Conference on Applications of Computer Vision (WACV)}, pages 3952--3961, 2021.

\bibitem{STBP}
Yujie Wu, Lei Deng, Guoqi Li, Jun Zhu, and Luping Shi.
\newblock Spatio-temporal backpropagation for training high-performance spiking neural networks.
\newblock {\em Frontiers in Neuroscience}, 12, 2018.

\bibitem{ponghiran_spiking_2022}
Wachirawit Ponghiran and Kaushik Roy.
\newblock Spiking {Neural} {Networks} with {Improved} {Inherent} {Recurrence} {Dynamics} for {Sequential} {Learning}.
\newblock In {\em Proceedings of the {AAAI} {Conference} on {Artificial} {Intelligence}}, pages 8001--8008, jun 2022.

\bibitem{Spikformer}
Zhaokun Zhou et~al.
\newblock Spikformer: When spiking neural network meets transformer.
\newblock In {\em The Eleventh International Conference on Learning Representations}, 2023.

\bibitem{ding2024shrinking}
Yongqi Ding, Lin Zuo, Mengmeng Jing, Pei He, and Yongjun Xiao.
\newblock Shrinking your timestep: Towards low-latency neuromorphic object recognition with spiking neural networks.
\newblock In {\em Proceedings of the {AAAI} {Conference} on {Artificial} {Intelligence}}, pages 11811--11819, 2024.

\bibitem{TKS}
Yiting Dong, Dongcheng Zhao, and Yi~Zeng.
\newblock Temporal knowledge sharing enable spiking neural network learning from past and future.
\newblock {\em IEEE Transactions on Artificial Intelligence}, 2024.

\bibitem{9412147}
Ravi~Kumar Kushawaha, Saurabh Kumar, Biplab Banerjee, and Rajbabu Velmurugan.
\newblock Distilling spikes: Knowledge distillation in spiking neural networks.
\newblock In {\em 2020 25th International Conference on Pattern Recognition (ICPR)}, pages 4536--4543, 2021.

\bibitem{9410323}
Sugahara Takuya, Renyuan Zhang, and Yasuhiko Nakashima.
\newblock Training low-latency spiking neural network through knowledge distillation.
\newblock In {\em 2021 IEEE Symposium in Low-Power and High-Speed Chips (COOL CHIPS)}, pages 1--3, 2021.

\bibitem{lv2023spikebert}
Changze Lv, Tianlong Li, Jianhan Xu, Chenxi Gu, Zixuan Ling, Cenyuan Zhang, Xiaoqing Zheng, and Xuanjing Huang.
\newblock Spikebert: A language spikformer trained with two-stage knowledge distillation from bert, 2023.

\bibitem{bal2023spikingbert}
Malyaban Bal and Abhronil Sengupta.
\newblock Spikingbert: Distilling bert to train spiking language models using implicit differentiation, 2023.

\bibitem{hong2023lasnn}
Di~Hong, Jiangrong Shen, Yu~Qi, and Yueming Wang.
\newblock Lasnn: Layer-wise ann-to-snn distillation for effective and efficient training in deep spiking neural networks, 2023.

\bibitem{jointasnn}
Yufei Guo, Weihang Peng, Yuanpei Chen, Liwen Zhang, Xiaode Liu, Xuhui Huang, and Zhe Ma.
\newblock Joint a-snn: Joint training of artificial and spiking neural networks via self-distillation and weight factorization.
\newblock {\em Pattern Recognition}, 142:109639, 2023.

\bibitem{xu2023biologically}
Qi~Xu, Yaxin Li, Xuanye Fang, Jiangrong Shen, Jian~K. Liu, Huajin Tang, and Gang Pan.
\newblock Biologically inspired structure learning with reverse knowledge distillation for spiking neural networks, 2023.

\bibitem{8891809}
Emre~O. Neftci, Hesham Mostafa, and Friedemann Zenke.
\newblock Surrogate gradient learning in spiking neural networks: Bringing the power of gradient-based optimization to spiking neural networks.
\newblock {\em IEEE Signal Processing Magazine}, 36(6):51--63, 2019.

\bibitem{TET}
Shikuang Deng, Yuhang Li, Shanghang Zhang, and Shi Gu.
\newblock Temporal efficient training of spiking neural network via gradient re-weighting, 2022.

\bibitem{IMLoss}
Yufei Guo et~al.
\newblock Im-loss: Information maximization loss for spiking neural networks.
\newblock In S.~Koyejo, S.~Mohamed, A.~Agarwal, D.~Belgrave, K.~Cho, and A.~Oh, editors, {\em Advances in Neural Information Processing Systems}, volume~35, pages 156--166. Curran Associates, Inc., 2022.

\bibitem{RecDis}
Yufei Guo et~al.
\newblock Recdis-snn: Rectifying membrane potential distribution for directly training spiking neural networks.
\newblock In {\em Proceedings of the IEEE/CVF Conference on Computer Vision and Pattern Recognition (CVPR)}, pages 326--335, June 2022.

\bibitem{tdBN}
Hanle Zheng, Yujie Wu, Lei Deng, Yifan Hu, and Guoqi Li.
\newblock Going {Deeper} {With} {Directly}-{Trained} {Larger} {Spiking} {Neural} {Networks}.
\newblock {\em Proceedings of the AAAI Conference on Artificial Intelligence}, 35(12):11062--11070, May 2021.

\bibitem{TEBN}
Chaoteng Duan, Jianhao Ding, Shiyan Chen, Zhaofei Yu, and Tiejun Huang.
\newblock Temporal effective batch normalization in spiking neural networks.
\newblock In S.~Koyejo, S.~Mohamed, A.~Agarwal, D.~Belgrave, K.~Cho, and A.~Oh, editors, {\em Advances in Neural Information Processing Systems}, volume~35, pages 34377--34390. Curran Associates, Inc., 2022.

\bibitem{MPBN}
Yufei Guo, Yuhan Zhang, Yuanpei Chen, Weihang Peng, Xiaode Liu, Liwen Zhang, Xuhui Huang, and Zhe Ma.
\newblock Membrane potential batch normalization for spiking neural networks.
\newblock In {\em Proceedings of the IEEE/CVF International Conference on Computer Vision (ICCV)}, pages 19420--19430, October 2023.

\bibitem{li2023efficient}
Boyan Li, Luziwei Leng, Ran Cheng, Shuaijie Shen, Kaixuan Zhang, Jianguo Zhang, and Jianxing Liao.
\newblock Efficient deep spiking multi-layer perceptrons with multiplication-free inference, 2023.

\bibitem{autosnn}
Byunggook Na, Jisoo Mok, Seongsik Park, Dongjin Lee, Hyeokjun Choe, and Sungroh Yoon.
\newblock {A}uto{SNN}: Towards energy-efficient spiking neural networks.
\newblock In Kamalika Chaudhuri, Stefanie Jegelka, Le~Song, Csaba Szepesvari, Gang Niu, and Sivan Sabato, editors, {\em Proceedings of the 39th International Conference on Machine Learning}, volume 162 of {\em Proceedings of Machine Learning Research}, pages 16253--16269. PMLR, 17--23 Jul 2022.

\bibitem{PLIF}
Wei Fang, Zhaofei Yu, Yanqi Chen, Timoth\'ee Masquelier, Tiejun Huang, and Yonghong Tian.
\newblock Incorporating learnable membrane time constant to enhance learning of spiking neural networks.
\newblock In {\em Proceedings of the IEEE/CVF International Conference on Computer Vision (ICCV)}, pages 2661--2671, October 2021.

\bibitem{MLF}
Lang Feng, Qianhui Liu, Huajin Tang, De~Ma, and Gang Pan.
\newblock Multi-level firing with spiking ds-resnet: Enabling better and deeper directly-trained spiking neural networks.
\newblock In {\em Proceedings of the Thirty-First International Joint Conference on Artificial Intelligence}, pages 2471--2477, 7 2022.

\bibitem{BackEISNN}
Dongcheng Zhao, Yi~Zeng, and Yang Li.
\newblock Backeisnn: A deep spiking neural network with adaptive self-feedback and balanced excitatory–inhibitory neurons.
\newblock {\em Neural Networks}, 154:68--77, 2022.

\bibitem{GLIF}
Xingting Yao, Fanrong Li, Zitao Mo, and Jian Cheng.
\newblock {GLIF}: A unified gated leaky integrate-and-fire neuron for spiking neural networks.
\newblock In Alice~H. Oh, Alekh Agarwal, Danielle Belgrave, and Kyunghyun Cho, editors, {\em Advances in Neural Information Processing Systems}, 2022.

\bibitem{hinton2015distilling}
Geoffrey Hinton, Oriol Vinyals, and Jeff Dean.
\newblock Distilling the knowledge in a neural network, 2015.

\bibitem{MLD}
Ying Jin, Jiaqi Wang, and Dahua Lin.
\newblock Multi-level logit distillation.
\newblock In {\em 2023 IEEE/CVF Conference on Computer Vision and Pattern Recognition (CVPR)}, pages 24276--24285, 2023.

\bibitem{li2022curriculum}
Zheng Li, Xiang Li, Lingfeng Yang, Borui Zhao, Renjie Song, Lei Luo, Jun Li, and Jian Yang.
\newblock Curriculum temperature for knowledge distillation, 2022.

\bibitem{Yuan_2020_CVPR}
Li~Yuan, Francis~EH Tay, Guilin Li, Tao Wang, and Jiashi Feng.
\newblock Revisiting knowledge distillation via label smoothing regularization.
\newblock In {\em Proceedings of the IEEE/CVF Conference on Computer Vision and Pattern Recognition (CVPR)}, June 2020.

\bibitem{yuan2023studentfriendly}
Mengyang Yuan, Bo~Lang, and Fengnan Quan.
\newblock Student-friendly knowledge distillation, 2023.

\bibitem{yang2023knowledge}
Zhendong Yang, Ailing Zeng, Zhe Li, Tianke Zhang, Chun Yuan, and Yu~Li.
\newblock From knowledge distillation to self-knowledge distillation: A unified approach with normalized loss and customized soft labels, 2023.

\bibitem{zong2023better}
Martin Zong, Zengyu Qiu, Xinzhu Ma, Kunlin Yang, Chunya Liu, Jun Hou, Shuai Yi, and Wanli Ouyang.
\newblock Better teacher better student: Dynamic prior knowledge for knowledge distillation.
\newblock In {\em The Eleventh International Conference on Learning Representations}, 2023.

\bibitem{liu2023norm}
Xiaolong Liu, LUJUN LI, Chao Li, and Anbang Yao.
\newblock {NORM}: Knowledge distillation via n-to-one representation matching.
\newblock In {\em The Eleventh International Conference on Learning Representations}, 2023.

\bibitem{romero2015fitnets}
Adriana Romero, Nicolas Ballas, Samira~Ebrahimi Kahou, Antoine Chassang, Carlo Gatta, and Yoshua Bengio.
\newblock Fitnets: Hints for thin deep nets, 2015.

\bibitem{Heo_2019_ICCV}
Byeongho Heo, Jeesoo Kim, Sangdoo Yun, Hyojin Park, Nojun Kwak, and Jin~Young Choi.
\newblock A comprehensive overhaul of feature distillation.
\newblock In {\em Proceedings of the IEEE/CVF International Conference on Computer Vision (ICCV)}, October 2019.

\bibitem{Zipf}
Jiajun Liang, Linze Li, Zhaodong Bing, Borui Zhao, Yao Tang, Bo~Lin, and Haoqiang Fan.
\newblock Efficient one pass self-distillation with zipf's label smoothing.
\newblock In Shai Avidan, Gabriel Brostow, Moustapha Ciss{\'e}, Giovanni~Maria Farinella, and Tal Hassner, editors, {\em Computer Vision -- ECCV 2022}, pages 104--119, Cham, 2022. Springer Nature Switzerland.

\bibitem{BYOT}
Linfeng Zhang, Jiebo Song, Anni Gao, Jingwei Chen, Chenglong Bao, and Kaisheng Ma.
\newblock Be your own teacher: Improve the performance of convolutional neural networks via self distillation.
\newblock In {\em 2019 IEEE/CVF International Conference on Computer Vision (ICCV)}, pages 3712--3721, 2019.

\bibitem{10037455}
Thi~Diem Tran, Kien~Trung Le, and An~Luong~Truong Nguyen.
\newblock Training low-latency deep spiking neural networks with knowledge distillation and batch normalization through time.
\newblock In {\em 2022 5th International Conference on Computational Intelligence and Networks (CINE)}, pages 01--06, 2022.

\bibitem{VGG}
Karen Simonyan and Andrew Zisserman.
\newblock Very deep convolutional networks for large-scale image recognition, 2015.

\bibitem{ResNet}
Kaiming He, Xiangyu Zhang, Shaoqing Ren, and Jian Sun.
\newblock Deep residual learning for image recognition.
\newblock In {\em Proceedings of the IEEE Conference on Computer Vision and Pattern Recognition (CVPR)}, June 2016.

\bibitem{ijcai2021p362}
Taehyeon Kim, Jaehoon Oh, Nak~Yil Kim, Sangwook Cho, and Se-Young Yun.
\newblock Comparing kullback-leibler divergence and mean squared error loss in knowledge distillation.
\newblock In {\em Proceedings of the Thirtieth International Joint Conference on Artificial Intelligence}, pages 2628--2635, 8 2021.

\bibitem{BN}
Sergey Ioffe and Christian Szegedy.
\newblock Batch normalization: Accelerating deep network training by reducing internal covariate shift.
\newblock In {\em Proceedings of the 32nd International Conference on Machine Learning}, volume~37, pages 448--456. PMLR, 07--09 Jul 2015.

\bibitem{BranchyNet}
Surat Teerapittayanon, Bradley McDanel, and H.T. Kung.
\newblock Branchynet: Fast inference via early exiting from deep neural networks.
\newblock In {\em 2016 23rd International Conference on Pattern Recognition (ICPR)}, pages 2464--2469, 2016.

\bibitem{menon21a}
Aditya~K Menon, Ankit~Singh Rawat, Sashank Reddi, Seungyeon Kim, and Sanjiv Kumar.
\newblock A statistical perspective on distillation.
\newblock In Marina Meila and Tong Zhang, editors, {\em Proceedings of the 38th International Conference on Machine Learning}, volume 139 of {\em Proceedings of Machine Learning Research}, pages 7632--7642. PMLR, 18--24 Jul 2021.

\bibitem{Wang_2022_CVPR}
Xiao Wang, Haoqi Fan, Yuandong Tian, Daisuke Kihara, and Xinlei Chen.
\newblock On the importance of asymmetry for siamese representation learning.
\newblock In {\em Proceedings of the IEEE/CVF Conference on Computer Vision and Pattern Recognition (CVPR)}, pages 16570--16579, June 2022.

\bibitem{CIFAR100}
Alex Krizhevsky, Geoffrey Hinton, et~al.
\newblock Learning multiple layers of features from tiny images.
\newblock 2009.

\bibitem{ImageNet}
Jia Deng, Wei Dong, Richard Socher, Li-Jia Li, Kai Li, and Li~Fei-Fei.
\newblock Imagenet: A large-scale hierarchical image database.
\newblock In {\em 2009 IEEE conference on computer vision and pattern recognition}, pages 248--255. Ieee, 2009.

\bibitem{CIFAR10-DVS}
Hongmin Li, Hanchao Liu, Xiangyang Ji, Guoqi Li, and Luping Shi.
\newblock Cifar10-dvs: An event-stream dataset for object classification.
\newblock {\em Frontiers in Neuroscience}, 11, 2017.

\bibitem{DVS-Gesture}
Arnon Amir et~al.
\newblock A low power, fully event-based gesture recognition system.
\newblock In {\em 2017 IEEE Conference on Computer Vision and Pattern Recognition (CVPR)}, pages 7388--7397, 2017.

\bibitem{RMPloss}
Yufei Guo, Xiaode Liu, Yuanpei Chen, Liwen Zhang, Weihang Peng, Yuhan Zhang, Xuhui Huang, and Zhe Ma.
\newblock Rmp-loss: Regularizing membrane potential distribution for spiking neural networks.
\newblock In {\em Proceedings of the IEEE/CVF International Conference on Computer Vision (ICCV)}, pages 17391--17401, October 2023.

\bibitem{real_spike_2022}
Yufei Guo et~al.
\newblock Real {Spike}: {Learning} {Real}-{Valued} {Spikes} for {Spiking} {Neural} {Networks}.
\newblock In {\em Computer {Vision} – {ECCV} 2022}, volume 13672, pages 52--68. Springer Nature Switzerland, 2022.

\bibitem{SRP}
Zecheng Hao, Tong Bu, Jianhao Ding, Tiejun Huang, and Zhaofei Yu.
\newblock Reducing ann-snn conversion error through residual membrane potential.
\newblock In {\em Proceedings of the AAAI Conference on Artificial Intelligence}, volume~37, pages 11--21, 2023.

\bibitem{ren2023spiking}
Dayong Ren, Zhe Ma, Yuanpei Chen, Weihang Peng, Xiaode Liu, Yuhan Zhang, and Yufei Guo.
\newblock Spiking pointnet: Spiking neural networks for point clouds, 2023.

\bibitem{DSR}
Qingyan Meng, Mingqing Xiao, Shen Yan, Yisen Wang, Zhouchen Lin, and Zhi-Quan Luo.
\newblock Training high-performance low-latency spiking neural networks by differentiation on spike representation.
\newblock In {\em 2022 IEEE/CVF Conference on Computer Vision and Pattern Recognition (CVPR)}, pages 12434--12443, 2022.

\bibitem{yao2024spike}
Man Yao, Jiakui Hu, Zhaokun Zhou, Li~Yuan, Yonghong Tian, Bo~Xu, and Guoqi Li.
\newblock Spike-driven transformer.
\newblock {\em Advances in Neural Information Processing Systems}, 36, 2023.

\bibitem{lian23learnable}
Shuang Lian, Jiangrong Shen, Qianhui Liu, Ziming Wang, Rui Yan, and Huajin Tang.
\newblock Learnable surrogate gradient for direct training spiking neural networks.
\newblock In {\em Proceedings of the Thirty-Second International Joint Conference on Artificial Intelligence}, pages 3002--3010, 2023.

\bibitem{ASGL}
Ziming Wang, Runhao Jiang, Shuang Lian, Rui Yan, and Huajin Tang.
\newblock Adaptive smoothing gradient learning for spiking neural networks.
\newblock In {\em International Conference on Machine Learning}, pages 35798--35816. PMLR, 2023.

\bibitem{9746774}
Yi~Chen, Silin Zhang, Shiyu Ren, and Hong Qu.
\newblock Gradual surrogate gradient learning in deep spiking neural networks.
\newblock In {\em IEEE International Conference on Acoustics, Speech and Signal Processing (ICASSP)}, pages 8927--8931, 2022.

\bibitem{deng2023surrogate}
Shikuang Deng, Hao Lin, Yuhang Li, and Shi Gu.
\newblock Surrogate module learning: Reduce the gradient error accumulation in training spiking neural networks.
\newblock In {\em International Conference on Machine Learning}, pages 7645--7657. PMLR, 2023.

\bibitem{fang2021deep}
Wei Fang, Zhaofei Yu, Yanqi Chen, Tiejun Huang, Timoth{\'e}e Masquelier, and Yonghong Tian.
\newblock Deep residual learning in spiking neural networks.
\newblock {\em Advances in Neural Information Processing Systems}, 34:21056--21069, 2021.

\bibitem{zhang2024enhancing}
Yuhan Zhang, Xiaode Liu, Yuanpei Chen, Weihang Peng, Yufei Guo, Xuhui Huang, and Zhe Ma.
\newblock Enhancing representation of spiking neural networks via similarity-sensitive contrastive learning.
\newblock In {\em Proceedings of the AAAI Conference on Artificial Intelligence}, volume~38, pages 16926--16934, 2024.

\bibitem{SLTT}
Qingyan Meng, Mingqing Xiao, Shen Yan, Yisen Wang, Zhouchen Lin, and Zhi-Quan Luo.
\newblock Towards memory-and time-efficient backpropagation for training spiking neural networks.
\newblock In {\em Proceedings of the IEEE/CVF International Conference on Computer Vision}, pages 6166--6176, 2023.

\end{thebibliography}
}

\appendix

\newpage
\section{Appendix}

\subsection{Details of Experiments}
\label{ExperimentsDetail}

\subsubsection{Dataset}
\label{DatasetDetail}
We evaluated the proposed method on the static CIFAR10/100 and neuromorphic benchmark datasets CIFAR10-DVS and DVS-Gesture.

\textbf{CIFAR10/100:} CIFAR10/100 \cite{CIFAR100} includes 10/100 different classes of object images. The dataset contains 60,000 images of size $32\times32$, of which 50,000 are used for training and 10,000 for testing. For CIFAR10/100, we applied random horizontal flips and random cropping without any additional data augmentation.

\textbf{ImageNet~\cite{ImageNet}:} ImageNet is the challenging image recognition dataset with 1.28 million training images and 50k test images from 1000 object classes. We resized the images within it to $224 \times 224$ and applied standard data augmentation to process them, as in~\cite{yao2024spike}.

\textbf{CIFAR10-DVS:} CIFAR10-DVS \cite{CIFAR10-DVS} is the neuromorphic version of CIFAR10. There are 10,000 samples of event streams in CIFAR10-DVS with spatial size of $128\times128$ with 10 classes. Each event stream $x \in [t,x,y,p]$ indicates the change in pixel value or brightness at location $[x,y]$ at the moment $t$ relative to the previous moment. $p$ represents polarity, and positive polarity indicates an increase in pixel value or brightness, and vice versa.

For the preprocessing of the CIFAR10-DVS data, we used the same approach as in \cite{MLF}. The original event stream is split into multiple slices in 10ms increments, and each slice is integrated into a frame and downsampled to $42 \times 42$. CIFAR10-DVS is divided into training and test sets in the ratio of 9:1.

\textbf{DVS-Gesture:} DVS-Gesture \cite{DVS-Gesture} is a neuromorphic dataset for gesture recognition. DVS-Gesture contains a total of 11 event stream samples of gestures, 1176 for training and 288 for testing, with a spatial size of $128 \times 128$ for each sample. For DVS-Gesture data, we integrate the event stream into frames in 30ms increments and downsample to $32\times32$.

\subsubsection{Implementation Details}
The experiments were conducted with the PyTorch package. All models were run on NVIDIA TITAN RTX with 100 epochs of training. The initial learning rate is set to 0.1 and decays to one-tenth of the previous rate every 30 epochs. The batch size is set to 64. The stochastic gradient descent optimizer was used with momentum set to 0.9. The weight decay was set to 1e-4 and 1e-3 for the static CIFAR10/100 and neuromorphic datasets, respectively. For LIF neurons, set the membrane potential time constant $\tau = 2.0$ and the threshold $\vartheta = 1.0$. 

For the ImageNet dataset, we follow the training strategy of \cite{yao2024spike}. A Lamb optimizer was used to train 300 epochs with an initial learning rate of 0.0005 and adjust it with a cosine-decay learning rate (To reduce the training time, we only train the first 150 epochs). The batch size is set to 32.

\subsubsection{Network Structures}
In our experiments, we use both VGG-9 and ResNet-18 architectures to validate our method. The specific structure of these two networks and the way the stages are divided are shown in Table~\ref{model}. For ResNet-18, the spiking neurons after the addition operation in each residual block are moved in front of the addition operation, the same as in~\cite{MLF}.

\begin{table}[!h]
 \centering
 \begin{tabular}{ccc}
  \toprule
  Stage & VGG-9 & ResNet-18 \\
  \midrule
  1 & - & Conv($3 \times 3@64$)\\
  \hline
  2  &  \makecell{Conv($3 \times 3$@64) \\ Conv($3 \times 3$@128)} &
  \makecell{
  $\left(
 	    \begin{array}{cc}  
 			 \makecell{\text{Conv}(3 \times 3@64) \\ \text{Conv}(3 \times 3@64)}
        \end{array}
    \right)\times 2
  $}\\
  \hline
    & average pool(stride=2) & -\\
  \hline
  3  & \makecell{Conv($3 \times 3$@256) \\ Conv($3 \times 3$@256)} & 
  $\left(
 	    \begin{array}{cc}  
 			 \makecell{\text{Conv}(3 \times 3@128) \\ \text{Conv}(3 \times 3@128)}
        \end{array}
    \right)\times 2
  $\\
  \hline
    & average pool(stride=2) & -\\
  \hline
  4  & \makecell{Conv($3 \times 3$@512) \\ Conv($3 \times 3$@512)} & $\left(
 	\begin{array}{cc}  
 			 \makecell{\text{Conv}(3 \times 3@256) \\ \text{Conv}(3 \times 3@256)}
 \end{array}
 \right)\times 2$\\
  \hline
    & average pool(stride=2) & -\\
  \hline
  5  & \makecell{Conv($3 \times 3$@512) \\ Conv($3 \times 3$@512)} & $\left(
 	\begin{array}{cc}  
 			 \makecell{\text{Conv}(3 \times 3@512) \\ \text{Conv}(3 \times 3@512)}
 \end{array}
 \right)\times 2$\\
  \hline
    \multicolumn{3}{c}{global average pool, fc}\\
  \bottomrule
 \end{tabular}
  \caption{Structures of VGG-9 and ResNet-18, where fc denotes the fully connected layer.}
 \label{model}
\end{table}

\subsubsection{Details of Reproduction of Existing Methods}
\label{self-implemented}
For reproducing the existing methods, the network structure, hyperparameters, and training method are the same as our TSSD, if not otherwise specified.

\textbf{teacher default-KD:} We implement teacher default-KD \cite{xu2023biologically} using the same architecture, parameters, and training strategy as our TSSD. For the default teacher signal, we set the class corresponding to the target to 0.95, with equal values for the remaining classes.

\textbf{Spike-driven Transformer:} For the reproduction of Spike-driven Transformer~\cite{yao2024spike} we use directly the official code of the paper. To avoid the negative effects of data differences, we integrated the DVS-Gesture inputs of Spike-driven Transformer at 30 ms intervals, the same as the inputs of our TSSD method.

\textbf{ASGL:} We reproduce the ASGL method using the training strategy from the original paper \cite{ASGL}. For a fair comparison, we processed the neuromorphic dataset using the data processing described in \ref{DatasetDetail} and evaluated the performance of ASGL. The experiments on DVS-Gesture have the same training strategy as CIFAR10-DVS, and we directly use the code provided by ASGL to modify only the dataset.

\textbf{PLIF:} We set the membrane potential time constant $\tau$ of LIF neurons in PLIF \cite{PLIF} as a learnable parameter, with each layer of neurons having the same $\tau$. Training starts with an initial $\tau$ value of 2.0.

\textbf{MLF:} We replaced the LIF neurons in the VGG-9 network with three levels of MLF \cite{MLF} neurons with firing thresholds of 0.6, 1.6, and 2.6, respectively, and kept the rest of the parameters and structure unchanged.

\textbf{TEBN:} We replaced the Batch Normalization layer in the VGG-9 network with TEBN \cite{TEBN}, leaving the rest of the parameters and structure unchanged.

\textbf{Spikformer:} In reproducing Spikformer \cite{Spikformer}, we use the same approach as in the original paper, i.e., data augmentation of the CIFAR10-DVS, using data of $128 \times 128$ size without downsampling directly into Spikformer. The loss function, learning rate adjustment policy, and batch size employed in the original paper are used. For DVS-Gesture, the data augmentation is not used. Training a total of 106 epochs on the neuromorphic datasets with Spikformer was the same configuration as the original paper on  CIFAR10-DVS.

\subsection{Accuracy Change Curves}
\label{acc}
\begin{figure}[h]
\centering
	\includegraphics[width=0.45\textwidth]{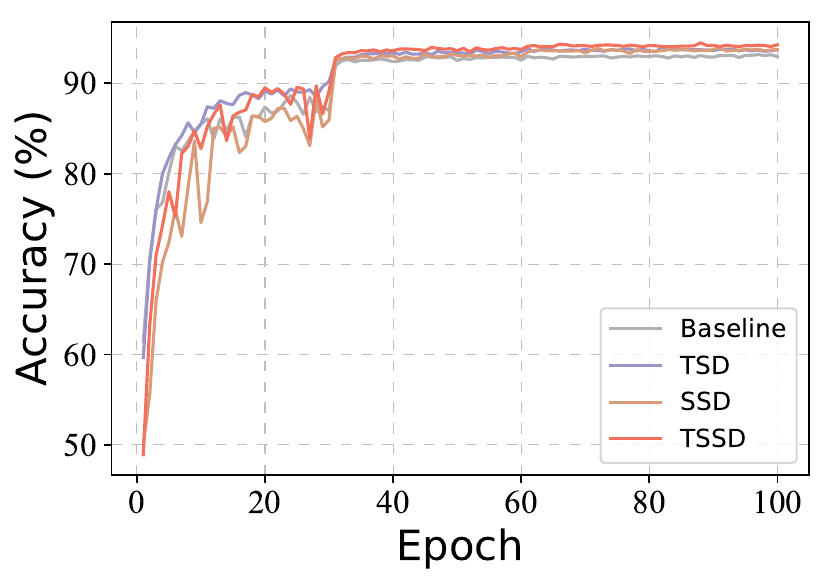}
	\includegraphics[width=0.45\textwidth]{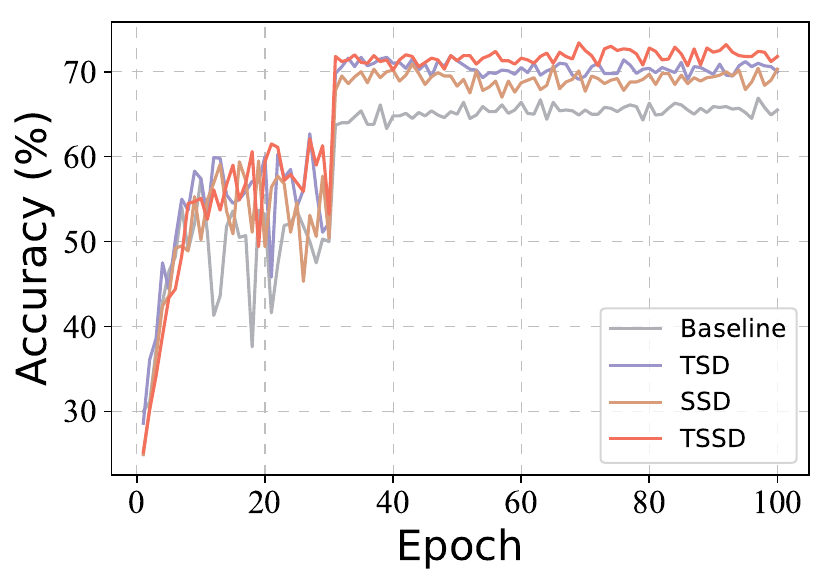}
\caption{Accuracy curves during training (left: CIFAR10, right: CIFAR10-DVS). }
\label{accuracy}
\end{figure}

To better illustrate the difference between the proposed method and the baseline, the accuracy change curves of the VGG network during training are shown in Fig.~\ref{accuracy}. The convergence accuracies of the proposed methods are consistently above the baseline, confirming the superior performance of our methods, a conclusion consistent with Table~\ref{tab:ablation}. It is worth noting that TSSD has a more stable accuracy curve than the baseline, suggesting that TSSD has a more stable training state, especially on CIFAR10-DVS.

\subsection{Spike Firing Rate Attention Map}
\label{sfram}
\begin{figure}[h]
\centering
	\includegraphics[width=0.99\textwidth]{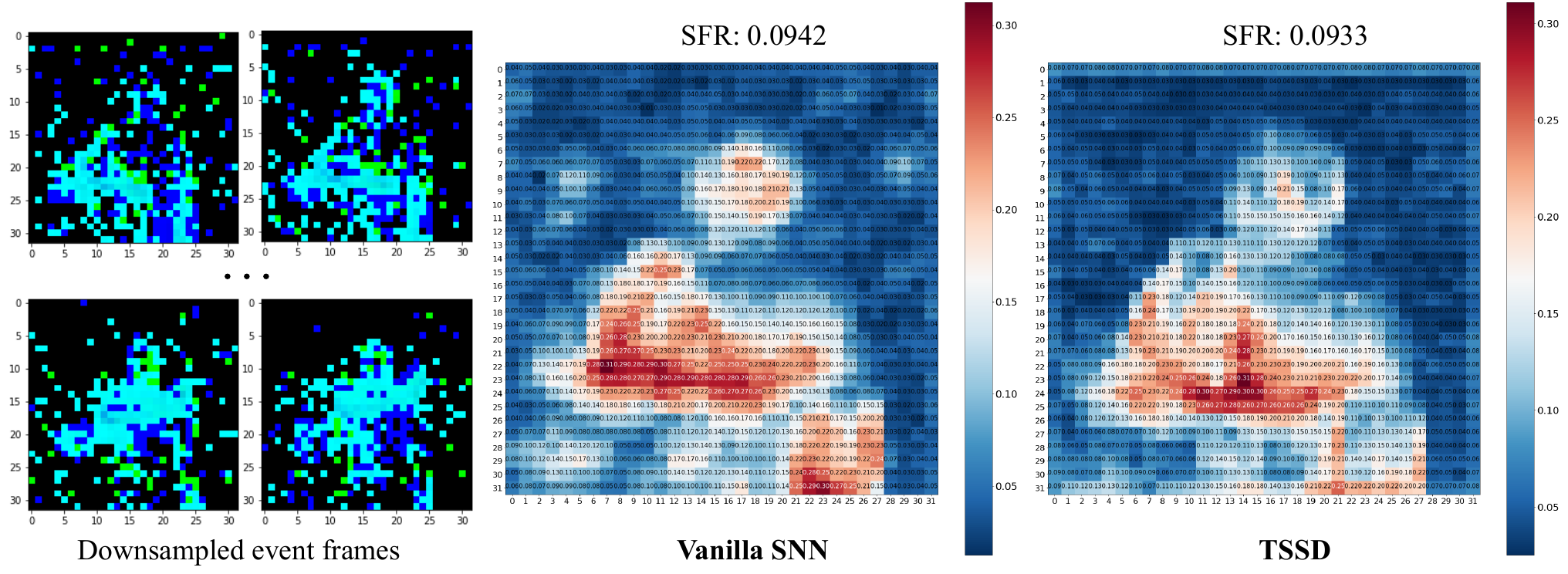}
\caption{Attention map based on spike firing rate (SFR). The attention areas of the vanilla SNN are more dispersed, while our TSSD focuses more on the hand region (lower left) which is more important for gesture recognition, and thus performs better.}
\label{sfr}
\end{figure}

The attention map based on spike firing rate on DVS-Gesture is plotted in Fig.~\ref{sfr} (To ensure visualization resolution, we plot the first convolution block in VGG-9). The most important area for gesture recognition is the waving hand to the lower left. In addition to the hand in the lower left, the attention area of the vanilla SNN covers part of the head area (upper center) and the lower right. These regions distract the vanilla SNN and therefore cause its limited performance. In contrast, our TSSD focuses more on the hand region and has a more concentrated attention, yielding higher recognition accuracy. In addition, our TSSD has a slightly lower SFR (0.0933) compared to the SFR of the vanilla SNN (0.0942) and is able to have lower energy consumption when deployed on neuromorphic hardware.

\subsection{Comparative Results on ImageNet}
The comparative results on ImageNet are shown in Table~\ref{com_imagenet}. To be fair, we compare our method with other methods under the same architecture. With 2 timesteps, our ResNet-34 achieves an accuracy of 66.13\%, outperforming Surrogate Module Learning (65.77\%)~\cite{deng2023surrogate} and SRP (64.32\%)~\cite{SRP} at the same latency. While SEW-ResNet (67.04\%)~\cite{fang2021deep} and Contrastive (66.78\%)~\cite{zhang2024enhancing} have higher performance, they have twice the latency of our method, and SEW-ResNet transmits integer information, which reduces its energy advantage. Notably, our method took only 150 epochs to train, which is less time consuming than these comparative methods. As training continues, TSSD is expected to deliver even better performance.

\begin{table*}[!h]
 \centering
 \caption{Comparative results on ImageNet.}
 \begin{threeparttable}
 \scalebox{0.9}{
 \begin{tabular}{cccccc}
  \toprule
 Method &Type & Architecture & $T$ & Spike form & Acc (\%)\\
  \midrule
  SLTT~\cite{SLTT} & Surrogate gradient & NF-ResNet-34 & 6 & Binary & 66.19\\
  TET~\cite{TET} & Surrogate gradient & ResNet-34 & 6 & Binary & 64.79\\
  STBP-tdBN~\cite{tdBN} & Surrogate gradient & ResNet-34 & 6 & Binary & 63.72\\
  SEW-ResNet~\cite{fang2021deep} & Surrogate gradient & SEW-ResNet-34 & 4 & \underline{Integer} & \underline{67.04}\\
  Contrastive~\cite{zhang2024enhancing}& Surrogate gradient & ResNet-34 & 4 & Binary & 66.78\\
  RMP-Loss~\cite{RMPloss} & Surrogate gradient & ResNet-34 & 4 & Binary & 65.17\\
  MPBN~\cite{MPBN} & Surrogate gradient & ResNet-34 & 4 & Binary & 64.31\\
  Surrogate Module~\cite{deng2023surrogate} & Surrogate Module & ResNet-34 & 2 & Binary & 65.77\\
  SRP~\cite{SRP} & Conversion & ResNet-34 & 2 & Binary & 64.32\\
  \hline
  \textbf{TSSD (Ours)} & Surrogate gradient+KD & ResNet-34 & 2 & Binary & \textbf{66.13}\\
  \bottomrule
 \end{tabular}
 }
 \end{threeparttable}
 \label{com_imagenet}
\end{table*}

\subsection{Weak Classifier Prediction Accelerated Inference}
\label{earlyexit}
Without compromising the inference efficiency of the SNN, we discard the weak classifier after training. However, if the weak classifiers are not discarded but used for prediction, it is possible to further improve the inference efficiency. To this end, we investigated the classification accuracy of the weak classifier in the hope of speeding up the inference process by recognizing simple samples using the weak classifier alone. The experimental results on CIFAR10-DVS, DVS-Gesture, and CIFAR100 are shown in Table~\ref{EarlyExit}. It can be seen that the recognition performance of the weak classifier is slightly degraded compared to the complete SNN, but it is still able to have satisfactory recognition accuracy (In particular, for DVS-Gesture, the recognition accuracy is reduced by only 2.78\%, but the spatial forward propagation path is halved.). The recognition of the weak classifier can be improved by further optimizing it or by training it on supervised signals. This makes it possible to use only the weak classifier to recognize simple samples during inference without forward passing features to the final classification layer, thus significantly reducing inference latency.

\begin{table}[t]
  \centering
  \caption{Comparative accuracy results of the complete SNN and the weak classifier (\%).}
  \scalebox{0.9}{
  \begin{tabular}{ccc}
    \toprule
    Dataset & Complete SNN & Weak classifier\\
    \hline
    CIFAR10-DVS & 72.70 & 64.50\\
    DVS-Gesture & 86.46 & 83.68\\
    CIFAR100 & 72.85 & 64.39\\
    \bottomrule
  \end{tabular}
  }
  \label{EarlyExit}
\end{table}

\end{document}